\documentclass{IET-Conf-Paper}

\usepackage[hidelinks]{hyperref}

\begin{document}

\title{On Input Formats for Radar Micro-Doppler Signature Processing by Convolutional Neural Networks}

\author{Mikolaj Czerkawski\ad{1}, Carmine Clemente\ad{1}, Craig Michie\ad{1}, Christos Tachtatzis\ad{1}}

\address{\add{1}{Department of Electronic and Electrical Engineering, University of Strathclyde, Glasgow, UK}
\email{mikolaj.czerkawski@strath.ac.uk}}


\begin{abstract}
    Convolutional neural networks have often been proposed for processing radar Micro-Doppler signatures, most commonly with the goal of classifying the signals. The majority of works tend to disregard phase information from the complex time-frequency representation. Here, the utility of the phase information, as well as the optimal format of the Doppler-time input for a convolutional neural network, is analysed. It is found that the performance achieved by convolutional neural network classifiers is heavily influenced by the type of input representation, even across formats with equivalent information. Furthermore, it is demonstrated that the phase component of the Doppler-time representation contains rich information useful for classification and that unwrapping the phase in the temporal dimension can improve the results compared to a magnitude-only solution, improving accuracy from 0.920 to 0.938 on the tested human activity dataset. Further improvement of 0.947 is achieved by training a linear classifier on embeddings from multiple-formats.
\end{abstract}

\maketitle

\section{Introduction}
\label{sec:introduction}

    Doppler-time representations are widely used for tasks involving high-level processing of Micro-Doppler radar signatures. Doppler-time representation is most commonly obtained via the Short-Time Fourier Transform of the rreceived radar signal, yielding 2D arrays of complex values. The spectrogram format is often extracted as the squared magnitude of the resulting complex STFT (Short-Time Fourier Transform). Despite the wide use of convolutional neural networks for processing this type of signals~\cite{Seyfioglu2018,Seyfioglu2019,Shrestha2020,Zhu2020,Addabbo2021,Li2022,CzerkawskiInterference,CzerkawskiCoherence}, it is not immediately obvious whether the spectrogram representation, or any other format, should be preferred. In this work, a set of experiments is designed and executed to provide further insight into the decision-making process in this context.
    
    The main contribution of this work is the analysis of the effect of input representation on the performance of the classifier, including a novel method based on multi-domain classifier. It has been found that the format of the input representation has a considerable influence on the performance of a convolutional neural network classifier, and hence, the findings presented here can bring a valuable impact on how classifiers of Micro-Doppler signatures are designed.
    
    A brief overview of the past approaches to the problem is provided in Section~\ref{sec:related_work}. This is followed by a description of the proposed experiments and discussion of the respective results in Section~\ref{sec:method}. The paper concludes with Section~\ref{sec:conclusion}.

\section{Related Work}
\label{sec:related_work}

    The focus of the presented work lies on the classification of targets from Doppler-time signal representation, a task, where each sample is mapped to a separate confidence score corresponding to an individual class. While there are other tasks that involve Doppler-time representations, such as image synthesis, the task of classification appears to be the most prominent and is hence the primary topic of this work.
    In the work of~\cite{Seyfioglu2018} introducing a deep convolutional autoencoder for radar-based human activity classification, the spectrograms are cropped to 4 seconds, and then converted to a single-channel representation of magnitude normalized to range of 0 and 1. A similar practice is followed in a related work on transfer learning techniques for Micro-Doppler motion classification in~\cite{Seyfioglu2019}, with further spatial downsampling employed. Further, in other works such as~\cite{Shrestha2020,Zhu2020,Addabbo2021,Li2022}, the authors operate on spectrogram representation. In a separate line of work, classifiers operating on real and imaginary 2-channel representations have been proposed in~\cite{CzerkawskiInterference,CzerkawskiCoherence}.

    To date, implementations of convolutional neural network models for classification of Doppler-time representations either rely on magnitude single-channel input, or real-imaginary double-channel. The experiments presented in this work aim to bridge the resulting gap, by exploring the utility of a phase-based representation and comparing it to the conventional formats.

\section{Methodology}
\label{sec:method}

    In this section, several approaches of supplying Doppler-time signals to convolutional neural networks are discussed, as well as potential methods of quantifying their effectiveness.

    \subsection{Potential Formats}
    \textbf{Polar 2-channel Format.} A commonly used representation of a Doppler-time power distribution is its magnitude. In some applications~\cite{Seyfioglu2018,Seyfioglu2019,Shrestha2020,Zhu2020,Addabbo2021,Li2022}, the polar magnitude component constitutes the entire representation. For completeness, the phases of the complex value can be maintained as a second channel. However, phase can be problematic due to its limited range of 2$\pi$ and the risk of wrapping, which can yield distortions in the signal. On the other hand, unwrapped phase can result in very large variance of the phase component, which could be difficult to process for a neural network. A potential solution could involve normalized unwrapped phase, but that changes the original phase scaling, which could be an important piece of information.
    
    \textbf{Rectangular 2-channel Format.} The issues with phase ambiguity can be alleviated by using the quadrature format, where real and imaginary parts of the complex value are mapped into individual channels. However, that makes the magnitude information harder to recover for the network (it must approximate $\ell_2$ norm, or learn other features that will deliver the equivalent information). Furthermore, the small differences in phase may be difficult to detect and again, a computationally complex inverse tangent operation must be approximated, or a set of features delivering equivalent information in a different manner.
    
    \textbf{Polar-Rectangular 4-channel Format.} An alternative solution is to use both rectangular and polar formats, in a 4-channel representation. For many applications, this should not greatly increase model size, since only the first (and potentially last) layers must be changed, doubling the number of weights compared to a 2-channel counterpart.
    
    
    \subsection{Processing Complex Data by Neural Networks}
        Deep Neural Networks are known to be highly expressive models, meaning that with enough width and depth of layers, they can approximate a wide set of functions. Based on this, it could be argued that the rectangular and polar representations can both be a valid format of the input data, since they contain equivalent and complete information. Should one format be favourable, the network is theoretically capable of learning an approximation of a function needed to transform into that format. For example, it is rather straightforward to learn the trigonometric function of sine over a finite range (which will indeed be finite in practical scenarios). However, learning these functions, or more complex $\ell_2$ norm required to calculate magnitude from rectangular format, may introduce an unnecessary burden on the network, where more capacity is spent on learning the transform, rather than the higher abstraction representations that are more useful for classification. This effect may be particularly apparent for scenarios with relatively small networks and small datasets, which is often the case for Doppler-time signal classification problems.
    
    \begin{figure}
        \centering
        \includegraphics[width=\columnwidth]{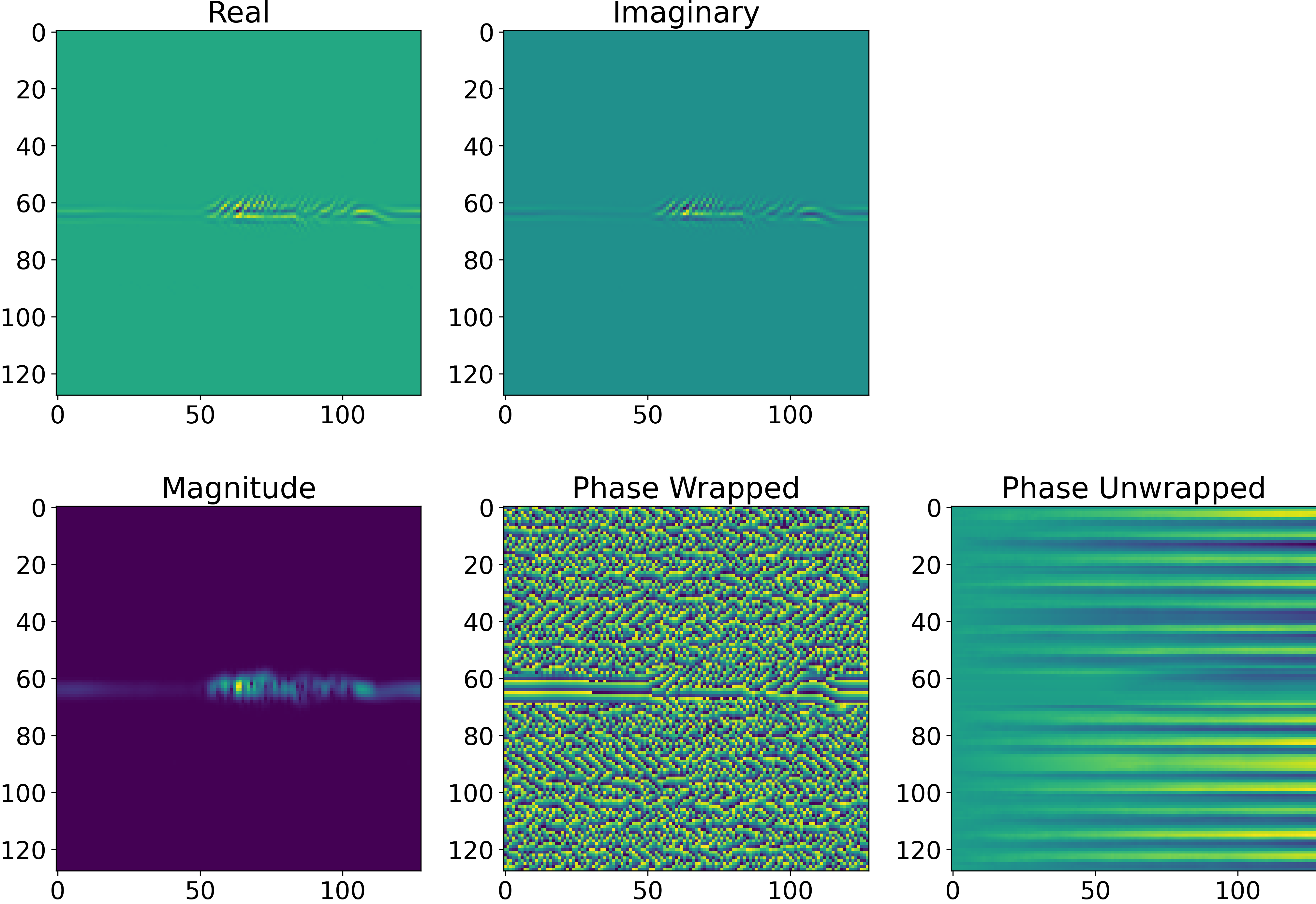}
        \caption{Single-channel formats considered in analysis.}
        \label{fig:formats}
    \end{figure}
    
    \subsection{Testing Methods}
    
        Two methods of testing the utility of each representation are proposed herein. First, a classification test with separate representation-specific models is conducted by training single-domain models on the same training dataset and with the same hyperparameters, and then comparing the achieved accuracy. However, the dynamics of each network training can be quite different and difficult to control. This motivates another type of experiment, where a model consisting of several domain-specific convolutional encoders is trained on all domain simultaneously (as later shown in Figure~\ref{fig:arch_multi}). To ensure that no domain input is ignored, at any step only 2 domain encodings are supplied to the final classifier head, selected at random, which forces the model to learn from all domains in order to minimize the classification error. This leads to a shared multi-domain model that is then analysed in several respects to provide insight into which input representations should be preferred.
    
\section{Results}
\label{sec:results}
    
    \subsection{Training Details}

    The dataset used for this work contains signatures of human motions (available at \url{http://researchdata.gla.ac.uk/848/})~\cite{glasgowdataset}, previously used in works such as~\cite{LiLSTM,LiTGRS,CzerkawskiInterference,CzerkawskiCoherence}. The dataset contains 1,752 samples of 6 human activities (walking, sitting Down, standing up, object pick up, drinking, and fall), and has been split into 50\% training, 25\% validation, and 25\% testing subsets. While each sample comes from the same FMCW radar that offers ranging capability, all range bins of the Range-Time representations are summed to yield a one-dimensional signature, that is then subject to Short-Time Fourier Transform with a 128-bin Blackman window of 0.2 seconds and 0.19-second overlap. The resulting time-frequency distribution is resampled to yield a 128 by 128 element matrix of complex values. Depending on the tested representation, the rectangular or polar components of the resulting complex value is obtained as the network input.
    
\begin{figure*}
        \centering
        \includegraphics[width=\textwidth]{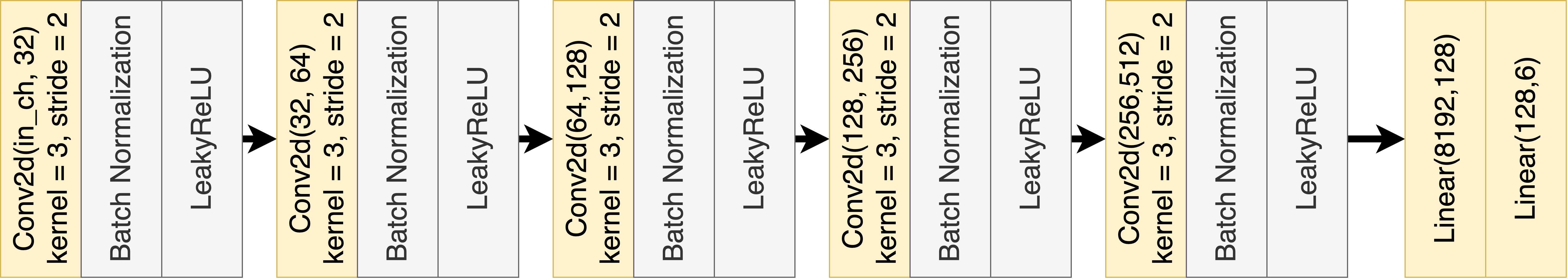}
        \caption{Diagram of the employed convolutional neural network architecture. The variable \textit{in\_ch} indicates the number of input channels that changes depending on the input representation format (either 1, 2, or 4 channels, as described in the text).}
        \label{fig:arch}
    \end{figure*}

    The employed convolutional neural network architecture is based on the model used in the previously published papers of~\cite{CzerkawskiInterference,CzerkawskiCoherence} (which are loosely based on models from~\cite{Seyfioglu2018,Seyfioglu2019}), accepting 128 by 128 image input, and it consists of 5 convolutional blocks, each consisting of a convolutional layer (kernel of 3 by 3 with a stride of 2), followed by a batch normalization layer and a LeakyReLU activation, as illustrated in Figure~\ref{fig:arch}. The output of the last convolutional block contains 8,192 values, and is linearly mapped to 128-dimensional intermediate embedding, which, in turn, is linearly mapped to 6 confidence scores. Depending on the type of the representation, the number of input channels \textit{in\_ch} of the 128 by 128 image is adjusted, as indicated in Figure~\ref{fig:arch}.
    
    The model is trained for a fixed number of 50~epochs with an Adam optimizer set to a learning rate of $10^{-4}$. A scheduler ReduceLROnPlateau with a factor of 0.5 and patience of 4 epochs is used to adjust the learning rate during the training process.
    
    \subsection{Classification Test (Single-Domain)}
    
        A direct way to compare the utility of each representation is to train the same model (apart from the rather minor difference in the number of input channels) with fixed hyperparameters and the number of epochs and compare the performance achieved with each input type. The results for such an experiment are contained in Table~\ref{tab:classification_performance}, where 10 different models have been trained on specific representations of interest, including a group of 5 polar-based representations (magnitude, phase (W), unwrapped phase (U)\footnote{For all experiments with phase unwrapping, this is executed in the temporal dimension, as shown in Figure~\ref{fig:formats}}, polar with raw phase (W), and polar with unwrapped phase (U)), a group of quadrature-based representations (real, imaginary, rectangular), and finally, a combined 4-channel representation containing rectangular and polar (with either wrapped or unwrapped phase). The accuracy is measured on the test set on both clean samples and samples with additive Gaussian noise of 0~dB SNR.
        
        \begin{table}[h]
            \caption{Classification accuracy with separate\\representation-specific models.}
            \centering
            \begin{tabular}{l c c}
                \toprule
                Mode & Accuracy & Accuracy \\
                 & (Clean) & 0 dB Noise \\
                \midrule
                Magnitude & 0.856 & 0.683 \\
                Phase (W)  & 0.237 & 0.224 \\
                Phase (U)  & 0.769  & 0.374 \\
                Polar 2-channel (W) & 0.349  & 0.269 \\
                Polar 2-channel (U) & 0.797  & 0.532 \\
                \midrule
                Real  & 0.820  & 0.744 \\
                Imaginary  & 0.833  & 0.756 \\
                Rectangular 2-channel  & 0.815  & 0.749 \\
                \midrule
                Pol-Rect 4-channel (W)  & 0.311  & 0.265\\
                Pol-Rect 4-channel (U)  & 0.795  & 0.441 \\
                \botrule
            \end{tabular}
            \label{tab:classification_performance}
        \end{table}
        
        As shown in Table~\ref{tab:classification_performance}, the highest level of performance is achieved using magnitude input (0.856), and quadrature-based representations (0.815-0.833). Out of these representations, the magnitude-based model appears to be more susceptible to noise (0.683 accuracy at 0dB) when compared to the quadrature-based models (0.744-0.756). Wrapped phase input yields very low performance of 0.237, yet its unwrapped equivalent achieves a far from trivial 0.769 accuracy. The disruptive nature of wrapped phase appears to also affect the performance when it is merely a part of the input representation, such as Polar 2-channel or Pol-Rect 4-channel. These observations verify the intuition that using the unwrapped phase leads to a significant boost of performance.
        
        At this point, an important finding worth noting is that phase information may contain useful discriminative features for Doppler-time signal classification, as indicated by the accuracy of 0.769 obtained with Phase (U). In this experiment, concatenation of magnitude with wrapped or unwrapped phase results in reduced performance compared to 0.856 accuracy achieved with magnitude-only input. This could be attributed to the relationship between the labels and the input being more complex than in the case of pure magnitude. This could be attributed to the fact that the magnitude and phase representations exhibit vastly different visual features and hence require disparate kernel filters. Since the same encoder topology is preserved (with the only change being the number of input channels), learning kernel weights that are optimal for the two diverse domains is challenging. One solution to this problem is presented in the following section, where both magnitude and phase representations are processed by representation-specific encoders to boost performance.
    
    \subsection{Classification Test (Multi-Domain)}
        
\begin{figure}
            \centering
            \includegraphics[width=\columnwidth]{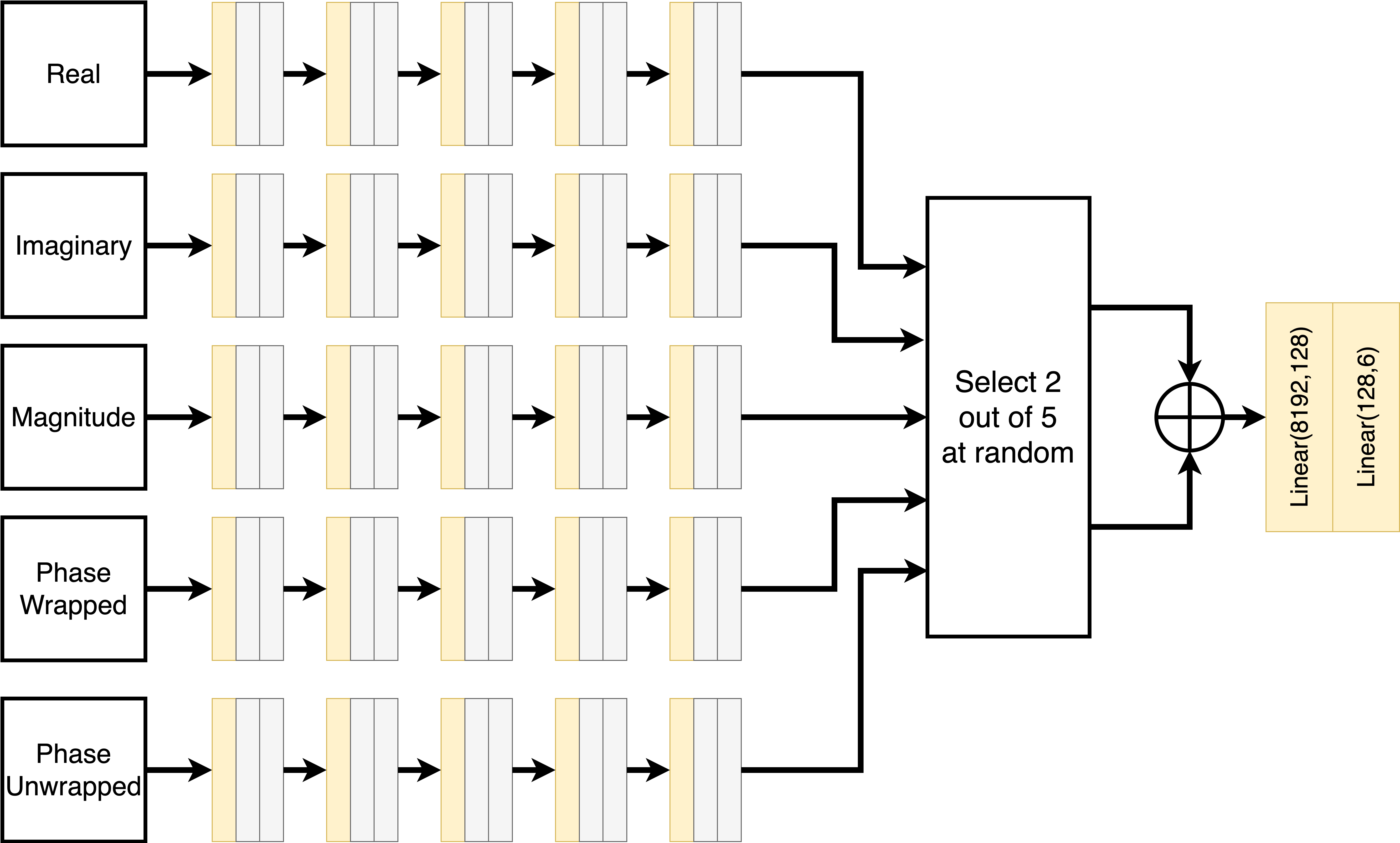}
            \caption{Multi-domain classifier training, where only 2 domains are used at any training step.}
            \label{fig:arch_multi}
        \end{figure}

        It is possible that each model reported in Table~\ref{tab:classification_performance} had a different training path due to a number of stochastic factors, like random sampling and network initialization. Testing with repetitions can mitigate that to some degree, but the space of variation can still remain large. An alternative is proposed to train a model accepting all 5 individual representations simultaneously with representation-specific convolutional modules (as illustrated in Figure~\ref{fig:arch_multi}) and summing the outputs of all modules to yield a combined representation. During training, the output of each entry module is omitted with uniform probability (by sampling only 2 out of 5 representations at any point), which aims to prevent the network from ignoring any individual representation. During testing, the accuracy achieved using any combination of representation can be compared as an indication of how important each one is. No scaling of the embedding vector is necessary, since the classifier head is linear, which guarantees preserved class confidence ratios in the output. The network is trained for 125 epochs instead of 50 to accommodate for the fact that each representation will be present in only 40\% of the optimization steps on average.
        
        Since the model has the flexibility to take any combination of the 5 trained representations, it is possible to test the level of performance achieved with many input format types, while reusing the same trained model. The resulting accuracy levels when utilizing each representation in isolation and combinations thereof are presented in Table~\ref{tab:classification_performance_combined}.
        
        In this case, it is shown that including unwrapped phase along with magnitude leads to a gain in accuracy from 0.920 to 0.938 under noiseless conditions. This type of representation, Polar 2-channel (U), leads to the highest accuracy of all representation in this experiment. Upon addition of 0dB Gaussian noise, the accuracy is still as high as 0.916, exceeded marginally only by the representation of Pol-Rect 4-channel (U), where 0.918 accuracy is reached. To further confirm the utility of phase information, it can be observed that the accuracy for the two phase-only representations is as high as 0.616 for wrapped phase and 0.877 for unwrapped phase. The rectangular representations seem to perform somewhat worse than polar-based counterparts, with an accuracy range of 0.893-0.904 under clean conditions. Lastly, using a combination of polar and rectangular does not seem to be beneficial when compared to the Polar 2-Channel (U) variant.

        \begin{table}[h]
            \caption{Classification performance with a combined\\Multi-Domain model.}
            \centering
            \begin{tabular}{l c c}
                \toprule
                Mode & Accuracy & Accuracy \\
                 & (Clean) & 0 dB Noise \\
                \midrule
                Magnitude & 0.920 & 0.863 \\
                Phase (W) & 0.616 & 0.461 \\
                Phase (U) & 0.877 & 0.760 \\
                Polar 2-channel (W) & 0.911 & 0.861 \\
                Polar 2-channel (U) & 0.938 & 0.916 \\
                \midrule
                Real & 0.893 & 0.856 \\
                Imaginary & 0.895 & 0.858\\
                Rectangular 2-channel & 0.904 & 0.893 \\
                \midrule
                Pol-Rect 4-channel (W) & 0.920 & 0.909 \\
                Pol-Rect 4-channel (U)& 0.925 & 0.918 \\
                Pol-Rect 5-channel (U\&W) & 0.927 & 0.906 \\
                \botrule
            \end{tabular}
            \label{tab:classification_performance_combined}
        \end{table}
        
        In both classification experiments, the wrapped phase representation appears to decrease performance when appended to another representation. This motivates another set of experiments aimed to explore further why and how the wrapped phase representation influences the classification process.
        
    \subsection{Further Analysis}
    
        First, it is explored whether there is any correlation between the classes predicted from individual representation. A strong correlation could mean that the test dataset contains samples that are inherently difficult to classify correctly based on what has been learned in the training dataset. To explore this, the incorrect predictions from each representation are compared to predictions based on the other representation, as shown in Table~\ref{tab:mode_correlation}. The first row contains the total number of incorrect predictions from each representation. The following table rows contain the fraction of these incorrect predictions that yield the same class prediction from another representation. For the problematic representation of wrapped phase, the consistency with other representations is quite small. For high-performing representations, like magnitude and imaginary, the overlap can be as high as 0.514, indicating that more than half of the incorrectly predicted samples based on the magnitude yields the same incorrect class label using imaginary representation, which could be a sign of inherently misguiding features in the training dataset that do not generalize well.
        
        \begin{table}[h]
            \caption{Comparison of achieved classification performance}
            \centering
            \begin{tabular}{l c c c c c}
                \toprule
                 Count &  47 & 46 & 35 & 168 & 54\\
                 & Re & Im & Mag & Ph (W) & Ph (U) \\
                \midrule
                Re & 1.000 & 0.478 & 0.429 & 0.065 & 0.278 \\
                Im & 0.468 & 1.000 & 0.514 & 0.054 & 0.204 \\
                Mag & 0.319 & 0.391 & 1.000 & 0.030 & 0.185 \\
                Ph (W) & 0.234 & 0.196 & 0.143 & 1.000 & 0.111 \\
                Ph (U) & 0.319 & 0.239 & 0.286 & 0.036 & 1.000 \\
                
                \botrule
            \end{tabular}
            \label{tab:mode_correlation}
        \end{table}
        
        An indication of the performance upper bound can be obtained by counting the number of samples where at least one modality leads to correct prediction. For the discussed experiment, this has been recorded to be the case for 432 samples out of the total 438 samples in the test dataset, yielding an upper bound in accuracy of 0.986. It means that potentially, even higher performance could be achieved by applying further processing to the output from individual representations (for example, internal embeddings or output confidences).
        
        Further perspective on this can be provided by counting the number of samples in the test dataset, where only a single representation yields the correct result. This is shown in Table~\ref{tab:unique}, where it can be seen that some representations do indeed yield a rather unique set of features that outperforms other representations, most prominently the unwrapped phase is the only representation correctly identifying 8 out of the 438 samples. In total, there is a total of 18 out of 438 samples that are correctly identified by only one of the representations, equivalent to a potential accuracy margin of 0.041. This means that the accuracy of 0.920 achieved using the magnitude representation could potentially be increased to 0.956 (+16/438) by taking advantage of unique predictions from other representations. This requires an additional processing step to pick-out the right set of predictions based on the multiple representations.
        
        \begin{table}[]
            \centering
            \caption{Number of unique samples in the test dataset where\\only one representation yields correct output.}
            \begin{tabular}{l r}
                \toprule
                Representation & Unique Sample Count \\
                \midrule
                Real & 0 out of 438 \\
                Imaginary & 4 out of 438 \\
                Magnitude & 2 out of 438 \\
                Phase (W) & 4 out of 438 \\
                Phase (U) & 8 out of 438 \\
                \midrule
                Total & 18 out of 438 \\
                \botrule
            \end{tabular}
            \label{tab:unique}
        \end{table}

        Whether this level of performance is achievable with the following encoders (or rather, how easy it is to apply a correction to a collection of mode-specific scores) is explored by training a `meta' classifier module (on the original training subset) that either takes the set of 5 embeddings of size 128 (provided by the penultimate layer in the diagram in Figure~\ref{fig:arch}), or the set of 6 final prediction scores. The extra model contains two hidden linear layers with 64 channels each, and is retrained for 10 epochs. Four configurations are tested, with one variable being the presence of LeakyReLU non-linearity in the hidden representations, and the other being the type of input (either a set of confidence scores of 5 x 6 or a set of embeddings of 5 x 128). The results in Table~\ref{tab:meta_acc} indicate that the top accuracy of 0.938 recorded for the polar representation with unwrapped phase can be pushed further to 0.947 by processing the embeddings with a linear module. For confidence-based `meta' module, no gain is observed.
        
        \begin{table}[]
            \centering
            \caption{Performance achieved by employing a 'meta'\\prediction module.}
            \begin{tabular}{llc}
                \toprule
                Input Type & Activation & Test Accuracy \\
                \midrule
                Confidence & Linear & 0.929 \\
                Confidence & LeakyReLU & 0.938 \\
                \midrule
                Embedding & Linear & 0.947 \\
                Embedding & LeakyReLU & 0.941 \\
                \midrule
                Base Model : Polar (U) & - & 0.938 \\
                \botrule
            \end{tabular}
            \label{tab:meta_acc}
        \end{table}
        
        \begin{figure}
            \centering
            \includegraphics[width=\columnwidth]{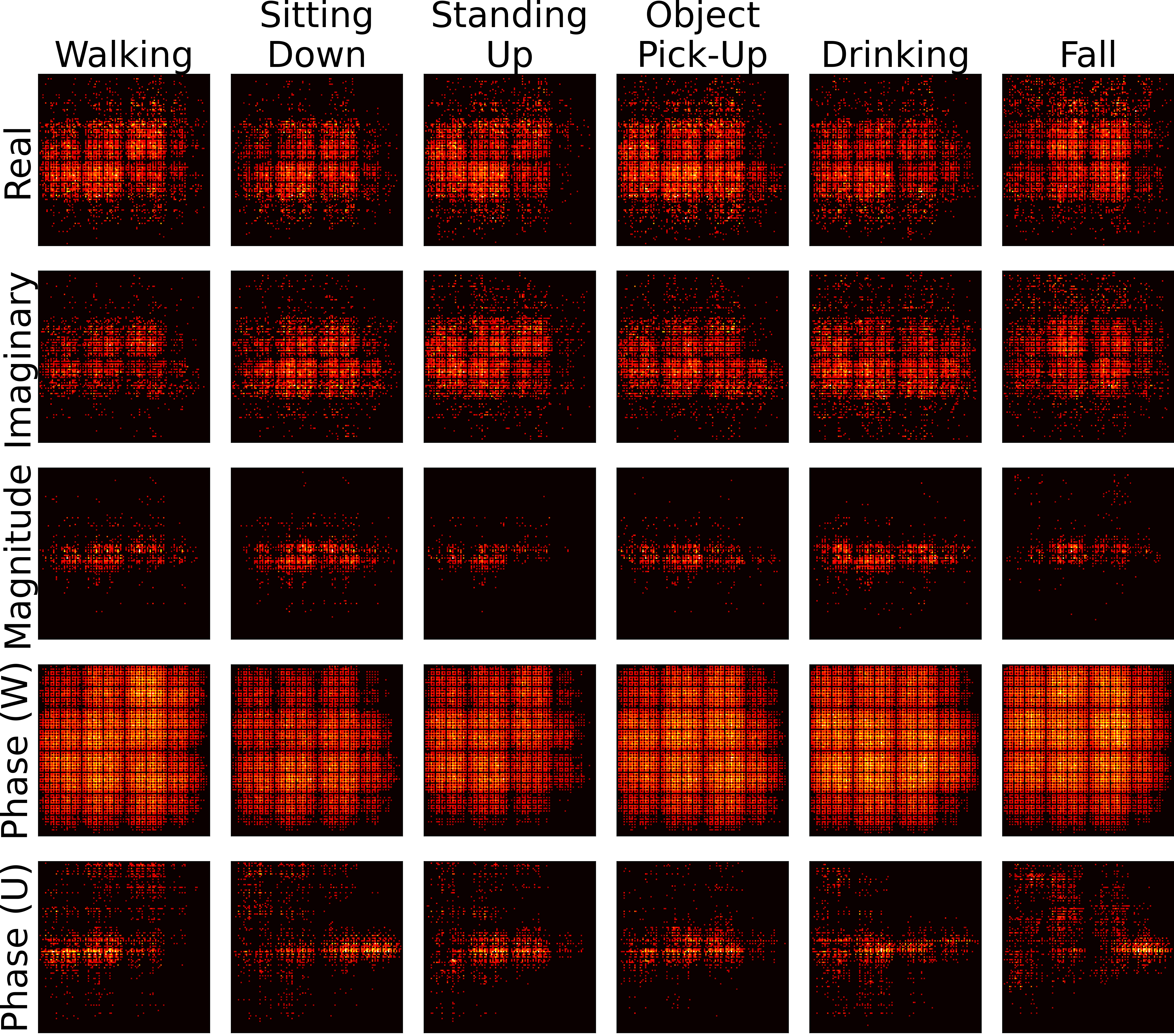}
            \caption{Illustration of saliency map for each input representation and each class. A threshold of 25\% of the maximum value is applied for visualization.}
            \label{fig:saliency}
        \end{figure}

        The phase unwrapping in temporal domain appears to bring significant benefits, while the wrapped phase is generally found to be a challenging representation to work with. This can be further confirmed by inspecting saliency maps~\cite{Simonyan2014} in Figure~\ref{fig:saliency} computed as the average gradient magnitude for each class on the test dataset. For visualization purposes, a threshold of 25\% of maximum value is applied. It can be observed that the gradients of the wrapped phase are distributed quite uniformly over the entire image, including high-velocity regions (top and bottom), where little information about the type of activity should be present. This effect is reduced in the case of unwrapped phase in the bottom row of the grid, similarly to the saliency maps obtained for the magnitude representation. The rectangular representation (Real and Imaginary) appear to also be quite sensitive to high-velocity regions, but not as strongly, and not as uniformly, as the wrapped phase, which is consistent with the achieved intermediate level of performance. In conclusion, it appears that the wrapped phase constitutes a noisy input representation, increasing the risk of spurious correlations in the training dataset that do not generalize well, and simply obscuring a degree of useful information. For this reason, even though the unwrapping operation may appear trivial, it has a strong effect on the capability of a convolutional neural network classifier.

\section{Conclusion}
\label{sec:conclusion}

    The presented work confirms the strong influence of Doppler-time representation format on convolutional neural network classifiers. The magnitude spectrogram representation remains the most forgiving input representation, significant information can be extracted from the phase component of the time-frequency distribution, especially when unwrapping in temporal domain is applied.
    
    Further analysis demonstrates that depending on the representation, different sets of features of the signatures are learned by a convolutional encoder, meaning that specific input representations can be more helpful than others for some of the samples. Further post-processing can be employed to take advantage of this divergence, as demonstrated herein.

    To the best of the authors' knowledge, this is the first systematic analysis of Doppler-time representations in the context of convolutional neural network classifiers. A novel method for assessing the utility of each representation type by training a Multi-Domain classifier is presented. The key impact of this work can be summarized by the improved performance from the conventional magnitude-based solution achieving 0.920, to 0.938 when unwrapped phase is included in the input representation, or to 0.947 when a post-processing meta module is employed.
    
    Further work could be done to confirm the same results on different datasets and different classification tasks that operate on radar Doppler-time signatures.


\end{document}